\documentclass{article}

\usepackage{microtype}
\usepackage{graphicx}
\usepackage{subcaption}
\usepackage{booktabs} 

\usepackage{hyperref}
\usepackage{xcolor}

\usepackage[accepted]{icml2019}

\usepackage{amsfonts}
\usepackage{amsmath}

\icmltitlerunning{Deep Reinforcement Learning with Decorrelation}

\begin{document}

\twocolumn[
\icmltitle{Deep Reinforcement Learning with Decorrelation}




\begin{icmlauthorlist}
\icmlauthor{Borislav Mavrin (borislav.mavrin@huawei.com)}{to,hu}
\icmlauthor{Hengshuai Yao (hengshuai.yao@huawei.com)}{si}
\icmlauthor{Linglong Kong (lkong@ualberta.ca)}{to,hu}
\end{icmlauthorlist}

\icmlaffiliation{to}{Department of Mathematical and Statistical Sciences, University of Alberta, Edmonton, Canada}
\icmlaffiliation{hu}{Huawei Noah's Ark}
\icmlaffiliation{si}{Huawei Hi-Silicon}

\icmlkeywords{Deep Reinforcement Learning}

\vskip 0.3in
]


\icmlcorrespondingauthor{Hengshuai Yao}{hengshuai.yao@huawei.com}

\printAffiliationsAndNotice{}  

\begin{abstract}
Learning an effective representation for high-dimensional data is a challenging problem in reinforcement learning (RL). Deep reinforcement learning (DRL) such as Deep Q networks (DQN) achieves remarkable success in computer games by learning deeply encoded representation from convolution networks. In this paper, we propose a simple yet very effective method for representation learning with DRL algorithms. Our key insight is that features learned by DRL algorithms are highly correlated, which interferes with learning. By adding a regularized loss that penalizes correlation in latent features (with only slight computation), we decorrelate features represented by deep neural networks incrementally. On 49 Atari games, with the same regularization factor, our decorrelation algorithms perform $70\%$ in terms of human-normalized scores, which is $40\%$ better than DQN. In particular, ours performs better than DQN on 39 games with 4 close ties and lost only slightly on $6$ games. Empirical results also show that the decorrelation method applies to Quantile Regression  DQN (QR-DQN) and significantly boosts performance. Further experiments on the losing games show that our decorelation algorithms can win over DQN and QR-DQN with a fined tuned regularization factor.
\end{abstract}

\section{Introduction} \label{sec:introduction}
Since the early days of artificial intelligence, learning effective representation of states has been one focus of research, especially because state representation is important for algorithms in practice. 
For example, reinforcement learning(RL)-based algorithms for computer Go used to rely on millions of binary features constructed from local shapes that recognize certain important geometry for the board. Some features were even constructed by the Go community from interviewing top Go professionals \cite{silver-thesis}. 
In this paper, we study state representation for RL. 
There has been numerous works in learning a good representation in RL with linear function approximation. 
Tile coding is a classical binary scheme for encoding states and generalizes in local sub-spaces \cite{cmac}. 
 \citet{analysis-lihong} constructed representation from sampled Bellman error to reduce policy evaluation error. 
\citet{Petrik:2007:ALM:1625275.1625690} computed powers of certain transition matrix in the underlying Markov Decision Processes (MDPs) to represent states. 
\citet{fourier} used Fourier basis functions to construct state representation; 
and so on. 

To the best of our knowledge, there are two approaches for state representation learning in DRL. 
The first is using auxiliary tasks. 
For example, UNREAL is an architecture that learns a universal representation by presenting the learning agent auxiliary tasks (each of which has a pseudo reward) that are relevant to the main task (which has an extrinsic reward) \cite{unreal}; with a goal of solving all the tasks from a shared representation. 
The second is a two-phase approach, which first learns a good representation and then performs learning and control using the representation.  The motivation for this line of work is to leverage neural networks for RL algorithms without using experience replay. 
It is well-known that neural networks has a ``catastrophic interference'' issue: the networks can forget what it has been trained on in the past, which is exactly the motivation for experience replay in DRL (using a buffer to store experience and replaying experience later). 
\citet{DBLP:journals/corr/abs-1805-07476} proposed to apply input transformation to reduce input interference caused by ReLU gates. 
\citet{DBLP:journals/corr/abs-1811-06626} proposed to learn sparse representation for features which can be used for online algorithms such as Sarsa.

While we also deal with representation learning with neural networks for RL, our motivation is to learn effective representation in a one-phase process, simultaneously acquiring a good representation and solving control. 
We also study the generic one-task setting, although our method may be applicable to multi-task control. 
Our insight in this paper is that \textbf{feature correlation} phenomenon happens frequently in DRL algorithms:
features learned by deep neural networks are highly correlated measured by covariances. 
We empirically show that the correlation causes DRL algorithms to have a slow learning curve. 
We propose to use a regularized correlation loss function to automatically decorrelate state features for Deep Q networks (DQN). 
Across 49 Atari games, our decorrelation method significantly improves DQN with only slight computation. 
Our decorrelation algorithms perform $70\%$ in terms of human-normalized scores and $40\%$ better than DQN. 
In particular, ours performs better than DQN on 39 games with 4 close ties with only slight loss on $6$ games. 

To test the generalibility of our method, we also apply decorrelation to Quantile Regression DQN (QR-DQN), a recent distributional RL algorithm that achieves generally better performance than DQN. Results show that the same conclusion holds: our decorrelation method performs much better than QR-DQN in terms of median human-normalized scores. 
In particular, the decorrelation algorithm lost 5 games, with 10 games being close (less than $5\%$ in cumulative rewards), and 34 winning games (more than $5\%$ in cumulative rewards). 
We used the same regularization factor in benchmarking for all games. 
We conducted further experiments on losing games, and found that the loss is due to a parameter choice for regularization factor. 
For the losing games, decorrelation algorithms with other values of regularization factor can significantly win over DQN and QR-DQN. 
We focused on discrete-action control in this paper, but similar techniques may be also applicable to continuous-action control, e.g., see \cite{zhang2018ace}.

\section{Background}\label{sec:background}
        We consider a Markov Decision Process (MDP) of a state space $\mathcal{S}$, an action space $\mathcal{A}$, a reward ``function'' $R: \mathcal{S} \times \mathcal{A} \rightarrow \mathbb{R}$, a transition kernel $p: \mathcal{S} \times \mathcal{A} \times \mathcal{S} \rightarrow [0, 1]$, and a discount ratio $\gamma \in [0, 1)$. In this paper we treat the reward ``function'' $R$ as a random variable to emphasize its stochasticity. Bandit setting is a special case of the general RL setting, where we usually only have one state.
        
        We use $\pi: \mathcal{S} \times \mathcal{A} \rightarrow [0, 1]$ to denote a stochastic policy. We use $Z^\pi(s, a)$ to denote the random variable of the sum of the discounted rewards in the future, following the policy $\pi$ and starting from the state $s$ and the action $a$. We have $Z^\pi(s, a) \doteq \sum_{t=0}^\infty \gamma^t R(S_t, A_t)$, where $S_0 = s, A_0 = a$ and $S_{t+1} \sim p(\cdot | S_t, A_t), A_t \sim \pi(\cdot| S_t)$. The expectation of the random variable $Z^\pi(s, a)$ is $$Q^\pi(s, a) \doteq \mathbb{E}_{\pi, p, R}[Z^\pi(s, a)]$$ which is usually called the state-action value function. 
        In general RL setting, we are usually interested in finding an optimal policy $\pi^*$, such that $Q^{\pi^*}(s, a) \geq Q^\pi(s, a)$ holds for any $(\pi, s, a)$. All the possible optimal policies share the same optimal state-action value function $Q^*$, which is the unique fixed point of the Bellman optimality operator \cite{bellman2013dynamic},
        \begin{align*}
        Q(s, a) = \mathcal{T}Q(s, a) \doteq \mathbb{E}[R(s, a)] + \gamma \mathbb{E}_{s^\prime \sim p}[\max_{a^\prime} Q(s^\prime, a^\prime)]
        \end{align*}
        Based on the Bellman optimality operator, \citet{watkins1992q} proposed Q-learning to learn the optimal state-action value function $Q^*$ for control. At each time step, we update $Q(s,a)$ as, 
        \begin{align*} 
        Q(s, a) \leftarrow Q(s, a) + \alpha (r + \gamma \max_{a^\prime}Q(s^\prime, a^\prime) - Q(s, a))
        \end{align*}
        where $\alpha$ is a step size and $(s, a, r, s^\prime)$ is a transition. There have been many work extending Q-learning to linear function approximation \cite{sutton2018reinforcement,szepesvari2010algorithms}. 

\newcommand{\real}{\mathbb{R}}

\section{Regularized Correlation Loss}\label{sec:body}

\subsection{DQN with Decorrelation}
\citet{mnih2015human} combined Q-learning with deep neural network function approximators, resulting the Deep-Q-Network (DQN). Assume the $Q$ function is parameterized by a network with weights $\theta$, at each time step, DQN performs a stochastic gradient descent to update $\theta$ minimizing the loss
        \begin{align*}
        \frac{1}{2}(r_{t+1} + \gamma \max_a Q_{\theta^-}(s_{t+1}, a) - Q_\theta(s_t, a_t)) ^ 2
        \end{align*}
        where $\theta^-$ is target network \cite{mnih2015human}, which is a copy of $\theta$ and is synchronized with $\theta$ periodically, and $(s_t, a_t, r_{t+1}, s_{t+1})$ is a transition sampled from a experience replay buffer \cite{mnih2015human}, which is a first-in-first-out queue storing previously experienced transitions.

Suppose the latent feature vector in the last hidden layer of DQN is denoted by a column vector $\phi \in \real^d$ (where $d$ is the number of units in the last hidden layer), then the covariance matrix of features estimated from samples is, 
\begin{equation}\label{eqn:correlation-matrix}
Var(\phi) = \frac{1}{n} \sum_{t=1}^n [\phi(s_t, a_t) - \bar{\phi}] [\phi(s_t, a_t) - \bar{\phi}]^T, 
\end{equation}
where $\bar{\phi} = \frac{1}{n} \sum_{t=1}^n \phi(s_t, a_t)$ and  $Var(\phi) \in \real^{d\times d}$.
We know that if two features $\phi_i$ and $\phi_j$ (the $i$th and $j$th unit in the last layer) are decorrelated, then $Var(\phi)_{i,j} = 0$. 
In linear function approximation, reinforcement learning algorithms ususally apply decorrelation to speed up learning via inverting some matrix that improves the conditioning of the underlying update equation, see e.g., \cite{ptd}. 

To apply decorrelation incrementally and efficiently for DQN, our key idea is to regularize the loss function with a term that penalizes correlation in features. 
        \begin{align*}
        \frac{1}{2}(r_{t+1} & + \gamma \max_a Q_{\theta^-}(s_{t+1}, a) - Q_\theta(s_t, a_t)) ^ 2 + \\ 
& \lambda  \frac{2}{d^2-d} \sum_{i > j}  \Big( Var(\phi)_{i,j} \Big)^2, 
        \end{align*}
where the regularization term is the mean-squared loss of the off-diagonal entries in the covariance matrix, which we call the {\em regularized correlation loss}.  
The other elements of our algorithm such as experience replay is completely the same as DQN. 
We call this new algorithm {\em DQN-decor}. 



\subsection{QR-DQN with Decorrelation}
        Instead of learning the expected return $Q$, distributional RL focuses on learning the full distribution of the random variable $Z$ directly \cite{jaquette1973markov,bellemare2017distributional,mavrin2019exploration}. 
Most distributional reinforcement learning algorithms use the distribution to estimate the mean of return. A recent work shows faster exploration can be achieved in both Atari and robot tasks by learning a policy over quantiles of the distribution, thus motivating distributional reinforcement learning beyond the mean-based return estimation \citep{zhang2018quota}. 

        The core idea behind QR-DQN is the use of Quantile Regression, e.g., see \cite{koenker1978regression,he2016regularized}. This approach gained significant attention in the field of theoretical and applied statistics. Let us first consider QR in the supervised learning. Given data $\{(x_i, y_i)\}_i$, we want to compute the quantile of $y$ corresponding the quantile level $\tau$.  Linear quantile regression loss is defined as:
        \begin{equation} \label{qr-loss}
            L(\beta) = \sum_i \rho_\tau(y_i - x_i \beta)
        \end{equation}
        where 
        \begin{equation} \label{check-function}
        \rho_\tau(u) = u (\tau - I_{u < 0}) = \tau |u| I_{u \ge 0} + (1 - \tau) |u| I_{u < 0}
        \end{equation}
        is the weighted sum of residuals. Weights are proportional to the counts of the residual signs and order of the estimated quantile $\tau$. For higher quantiles positive residuals get higher weight and vice versa.
        If $\tau=\frac{1}{2}$, then the estimate of the median for $y_i$ is $\theta_{1/2} (y_i|x_i) = x_i \hat{\beta}$, with $\hat{\beta} = \arg \min L(\beta)$.
        
        

There are various approaches to represent a distribution in RL setting \cite{bellemare2017distributional,dabney2018implicit,barth2018distributed}. In this paper, we focus on the quantile representation \cite{dabney2017distributional} used in QR-DQN, where the distribution of $Z$ is represented by a uniform mix of $N$ supporting quantiles:
        \begin{align*}
        Z_\theta(s, a) \doteq \frac{1}{N}\sum_{i=1}^N \delta_{\theta_i(s, a)}
        \end{align*}
        where $\delta_x$ denote a Dirac at $x \in \mathbb{R}$, and each $\theta_i$ is an estimation of the quantile corresponding to the quantile level (a.k.a. quantile index) $\tau_i \doteq \frac{i - 0.5}{N}$ for $0 < i \leq N$. The state-action value $Q(s, a)$ is then approximated by $\frac{1}{N}\sum_{i=1}^N \theta_i(s, a)$. Such approximation of a distribution is referred to as quantile approximation. 
        
        Similar to the Bellman optimality operator in mean-centered RL, we have the distributional Bellman optimality operator for control in distributional RL,
        \begin{align*}
        \mathcal{T}Z(s, a) \doteq R(s, a) + \gamma Z(s^\prime, \arg\max_{a^\prime}\mathbb{E}_{p, R}[Z(s^\prime, a^\prime)]) \\
        s^\prime \sim p(\cdot|s, a)
        \end{align*}
        Based on the distributional Bellman optimality operator, \citet{dabney2017distributional} proposed to train quantile estimations (i.e., $\{q_i\}$) via the Huber quantile regression loss \cite{huber1964robust}. To be more specific, at time step $t$ the loss is 
        \begin{align*}
        \frac{1}{N}\sum_{i=1}^N \sum_{i^\prime=1}^N\Big[\rho_{\tau_i}^\kappa\big(y_{t, i^\prime} - \theta_i(s_t, a_t)\big)\Big]
        \end{align*}
        where 
\begin{equation}\label{eqn:qrdqn-y}
y_{t, i^\prime} \doteq r_t + \gamma \theta_{i^\prime}\big(s_{t+1}, \arg\max_{a^\prime}\frac{1}{N}\sum_{i=1}^N \theta_i(s_{t+1}, a^\prime)\big)
\end{equation}
and 
        \begin{equation}\label{eqn:qrdqn-rho}
\rho_{\tau_i}^\kappa(x) \doteq |\tau_i - \mathbb{I}\{x < 0\}|\mathcal{L}_\kappa(x), 
\end{equation}
where $\mathbb{I}$ is the indicator function and $\mathcal{L}_\kappa$ is the Huber loss,
        \begin{align*}
            \mathcal{L}_\kappa(x) \doteq \begin{cases}
            \frac{1}{2}x^2 & \text{if } x \leq \kappa \\
            \kappa(|x| - \frac{1}{2}\kappa) & \text{otherwise}
            \end{cases}
        \end{align*}

To decorrelate QR-DQN, we use the following loss function, 
 \begin{align*}
        \frac{1}{N}\sum_{i=1}^N \sum_{i^\prime=1}^N & \Big[\rho_{\tau_i}^\kappa\big(y_{t, i^\prime} - \theta_i(s_t, a_t)\big)\Big] +  \\
  & \lambda  \frac{2}{d^2-d} \sum_{i > j}  \Big( Var(\phi)_{i,j} \Big)^2, 
        \end{align*}
where $\phi$ denotes the feature encoder by the last layer of QR-DQN, 
$y_{t, i^\prime}$ is computed in equation (\ref{eqn:qrdqn-y}), and $\rho$ is computed in equation (\ref{eqn:qrdqn-rho}). 

The other elements of QR-DQN are also preserved. The only difference of our new algorithm from QR-DQN is the regularized correlation loss. 
We call this new algorithm {\em QR-DQN-decor}. 



\section{Experiment}\label{sec:experiment}
In this section, we conduct experiments to study the effectiveness of the proposed decorrelation method for DQN and QR-DQN.
We used the Atari game environments by \citet{bellemare2013arcade}. In particular, we compare DQN-decor vs. DQN, and QR-DQN-decor vs. QR-DQN. Algorithms were evaluated on training with 20 million frames (or equivalently, 5 million agent steps) 3 runs per game.

In performing the comparisons, we used the same parameters for all the algorithms, which are reported below. 
The discount factor is $0.99$. The image size is $(84, 84)$. 
We clipped the continuous reward into three discrete values, $\{0,1,-1\}$, according to their sign. 
This clipping of reward signals leads to more stable performance according to \cite{mnih2015human}. 
The reported performance scores is calculated, however, in terms of the original unclipped rewards. 
The target networks update frequency is $10,000$ frames. 
The learning rate is $10^{-4}$. 
The exploration strategy is epsilon-greedy, with $\epsilon$ decaying linearly from $1.0$ to $0.02$ in the first 1 million frames, and remains constantly $0.02$ after that.  
The experience replay buffer size is 1 million. 
The minibatch size for experience replay is $32$.
In the first $200,000$ frames, all the four agents were behaving randomly to fill the buffer with experience as a warm start.

\begin{figure}[t]
\vskip 0.2in
\begin{center}
\centerline{\includegraphics[width=\columnwidth]{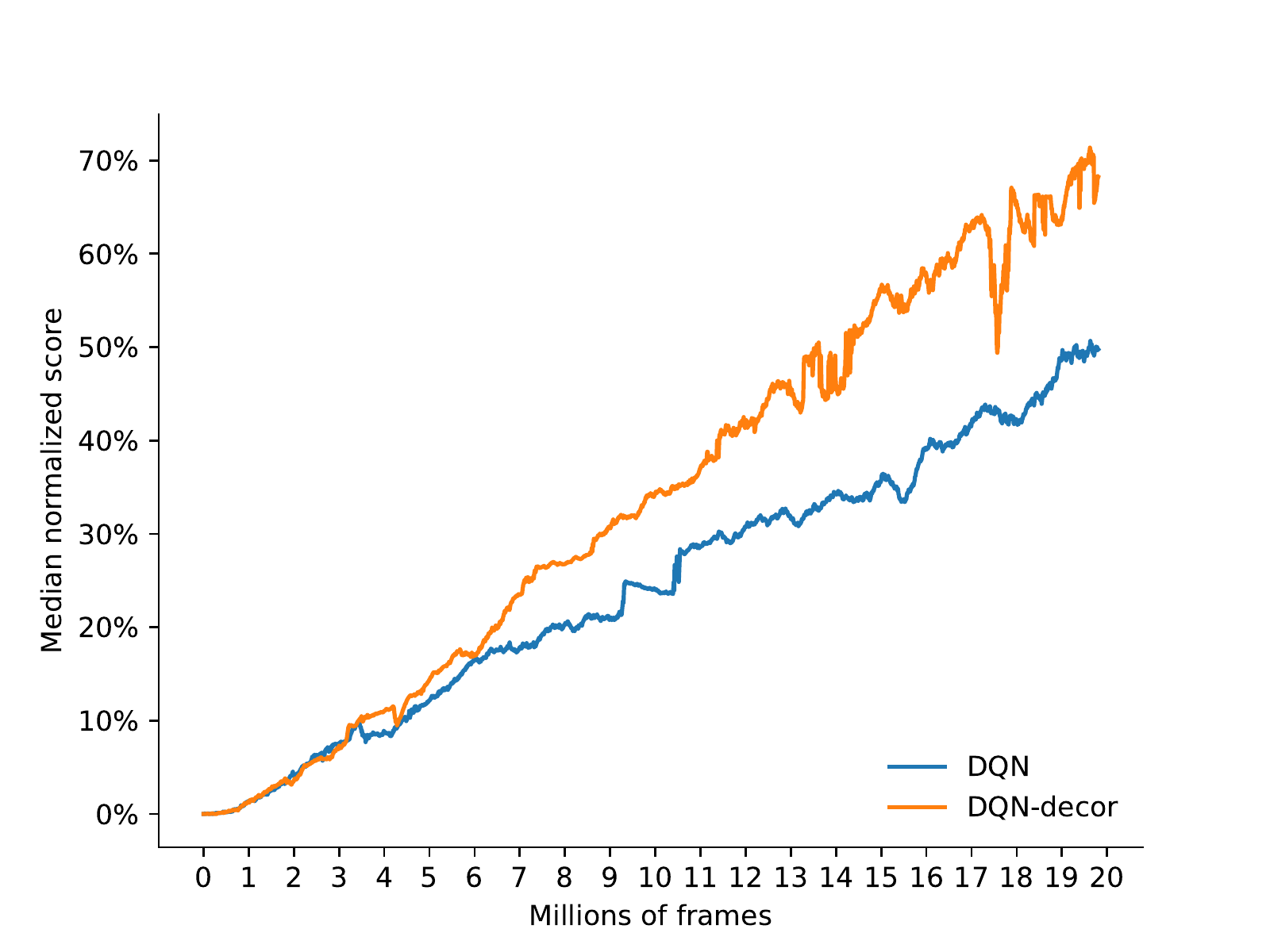}}
\caption{Human-normalized performance (using median across 49 Atari games): DQN-decor vs. DQN.}
\label{fig:dqn-corr-perf}
\end{center}
\vskip -0.2in
\end{figure}

\begin{figure}[t]
\vskip 0.2in
\begin{center}
\centerline{\includegraphics[width=\columnwidth]{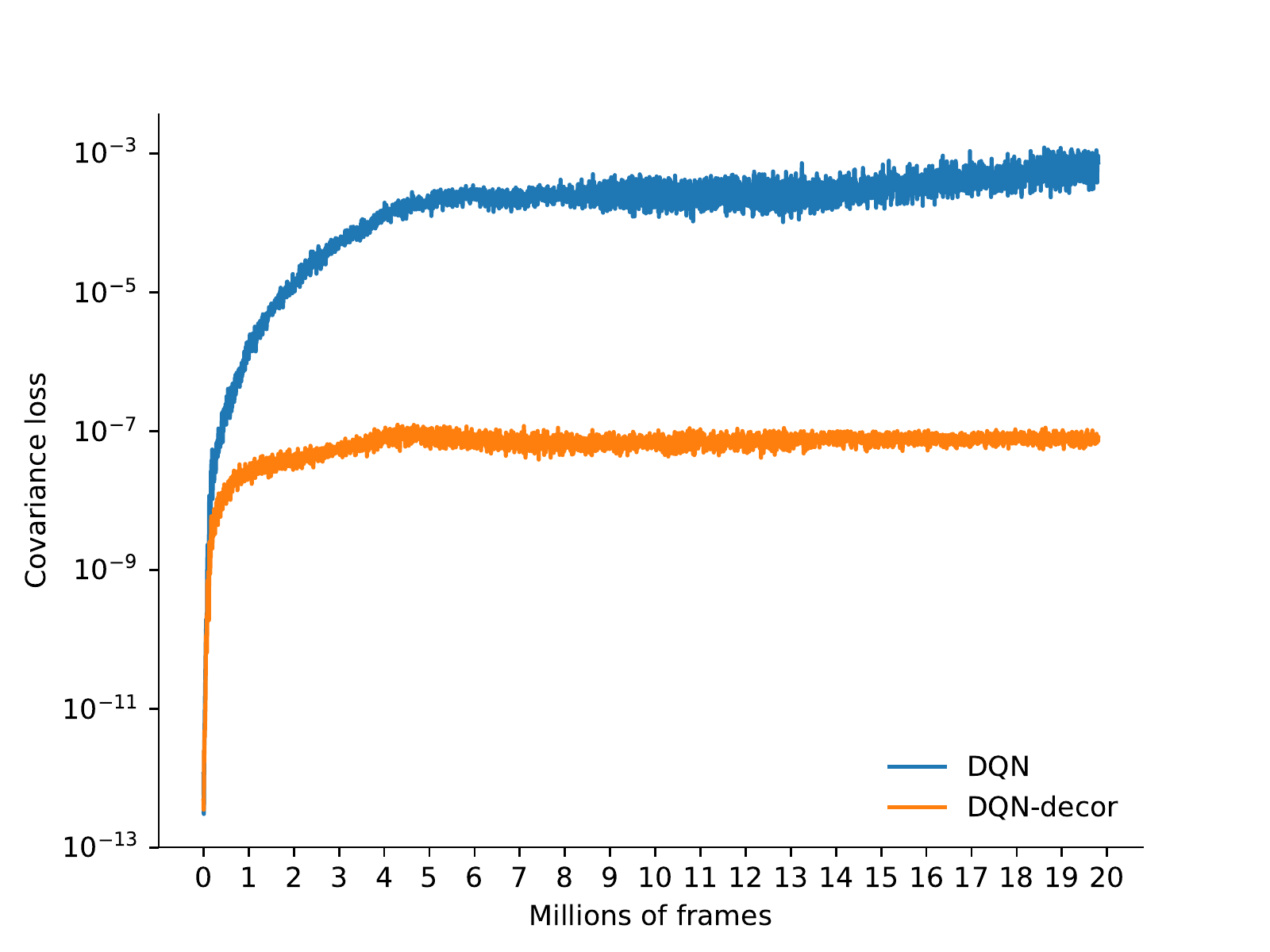}}
\caption{Correlation loss over training (using median across 49 Atari games): DQN-decor vs. DQN.}
\label{fig:dqn-corr-loss}
\end{center}
\vskip -0.2in
\end{figure}

\begin{figure*}[t]
\vskip 0.2in
\begin{center}
\centerline{\includegraphics[width=1.5\columnwidth]{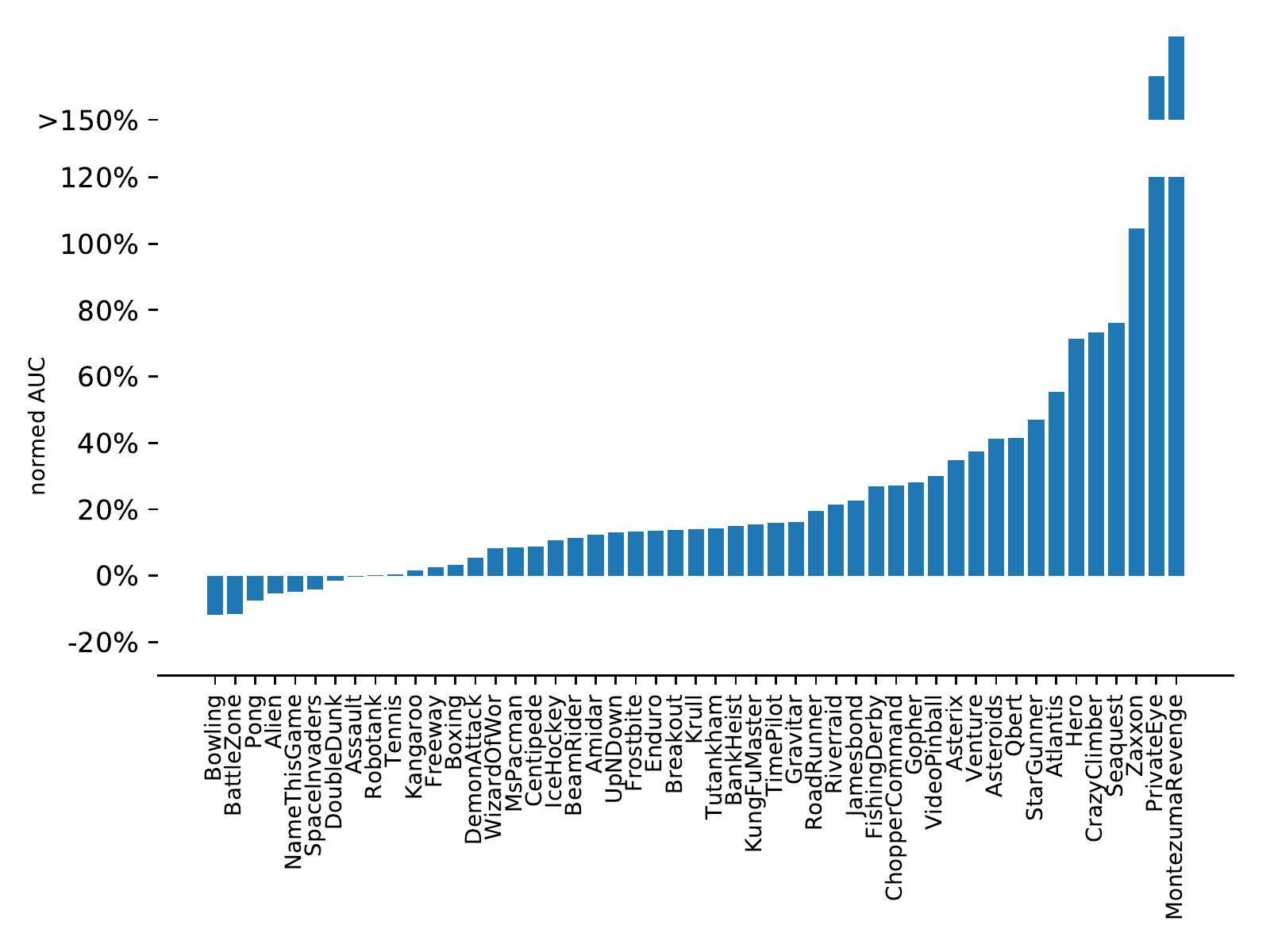}}
\caption{Cumulative reward improvement of DQN-decor over DQN. The bar indicates improvement and is computed as the  Area Under the Curve (AUC)  in the end of training, normalized relative to DQN. Bars above / below horizon indicate performance gain / loss. For MonteZumaRevenge, the improvement is $15133.34\%$; and for PrivateEye, the improvement is $334.87\%$. }
\label{fig:dqn-auc}
\end{center}
\vskip -0.2in
\end{figure*}

\begin{figure*}[t]
    \centering
    \begin{subfigure}[b]{0.33\linewidth}        
        \centering
        \includegraphics[width=0.7\linewidth]{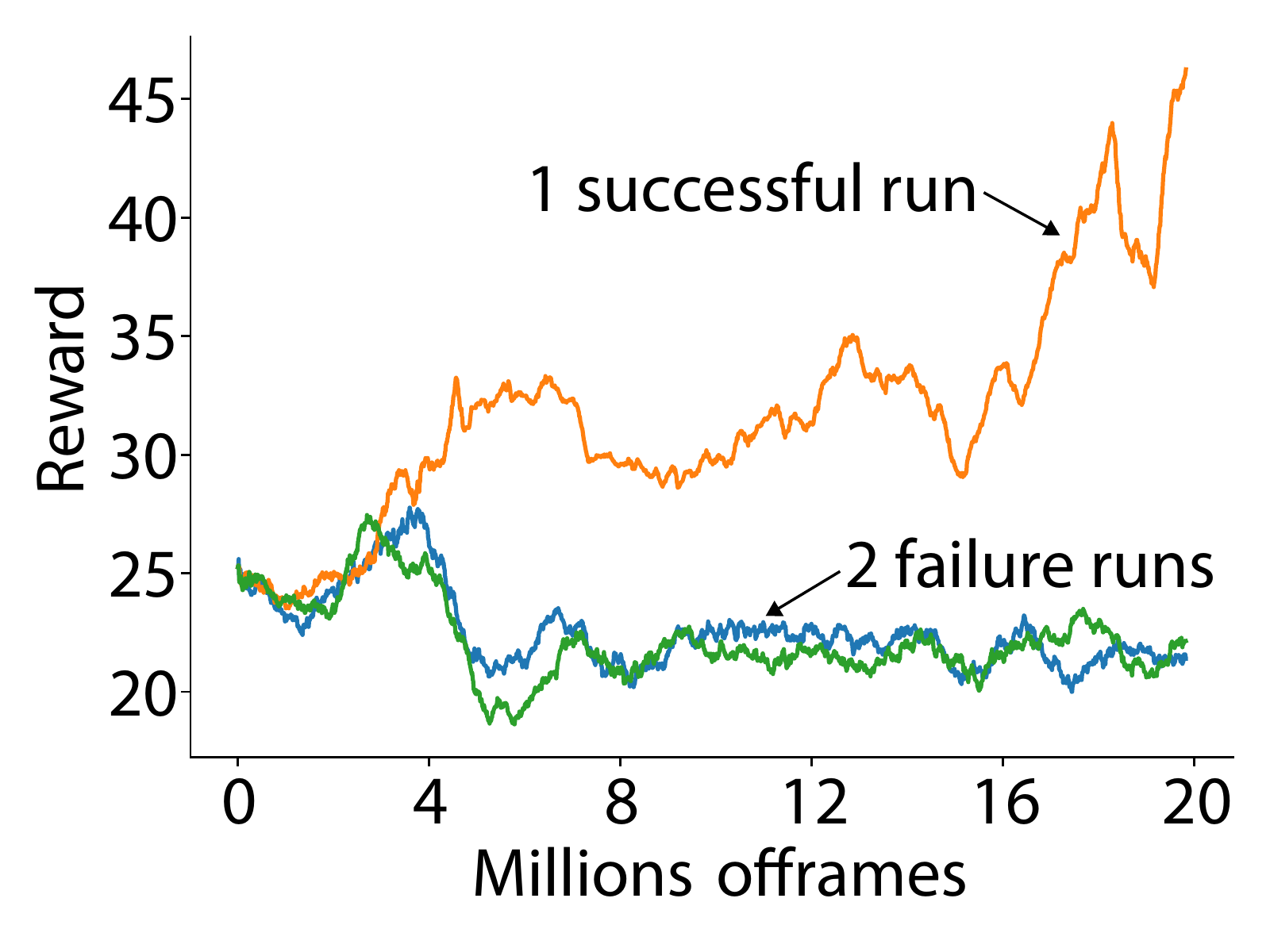}
        \caption{Bowling-DQN: Performance.}
        \label{fig:dqn-bowling-3runs}
    \end{subfigure}
    \begin{subfigure}[b]{0.33\linewidth}        
        \centering
        \includegraphics[width=0.7\linewidth]{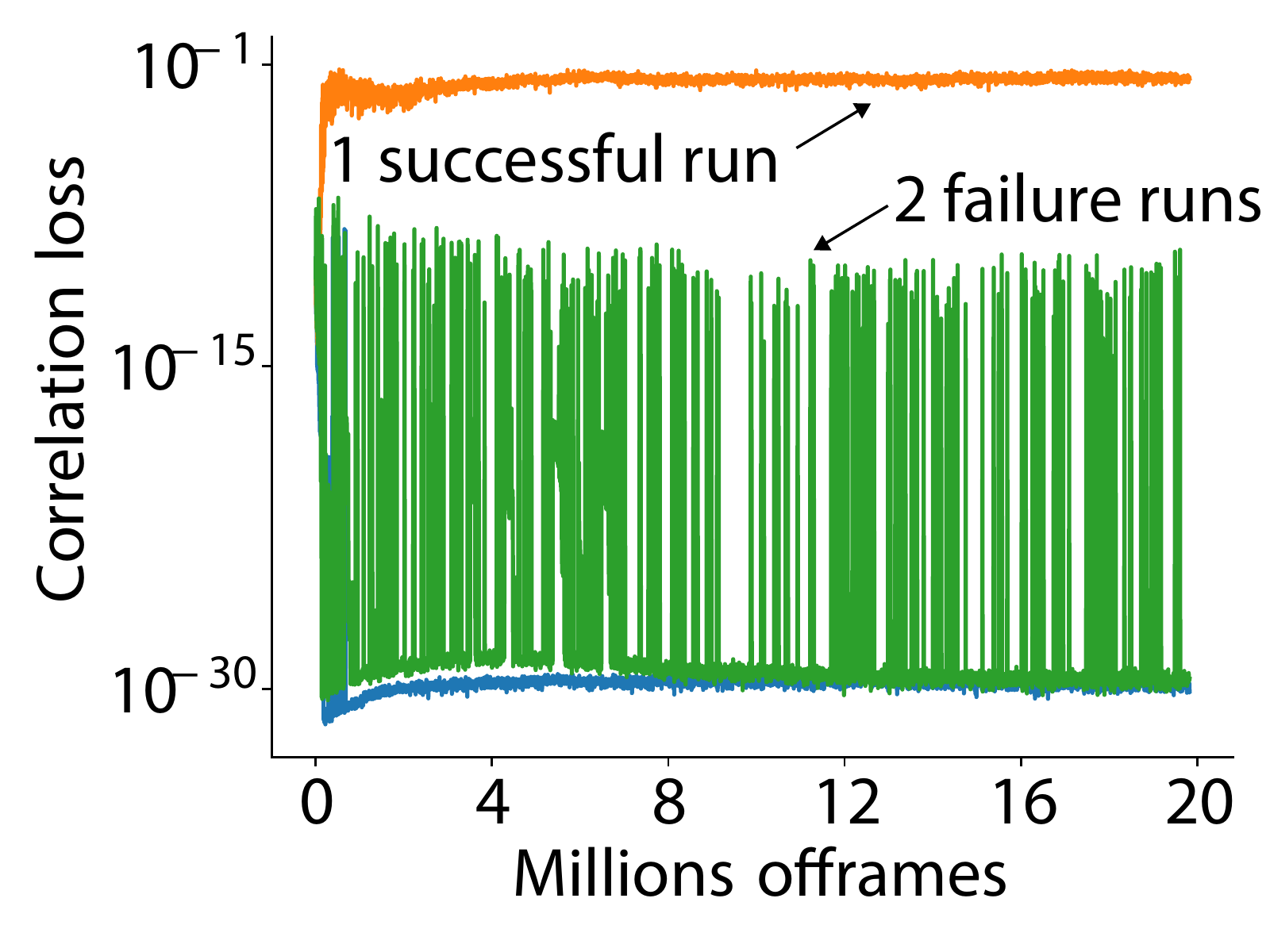}
        \caption{Bowling-DQN: Correlation loss. }
        \label{fig:dqn-bowling-3runs-cor}
    \end{subfigure}
\begin{subfigure}[b]{0.33\linewidth}
\centering
\centerline{\includegraphics[width=0.7\columnwidth]{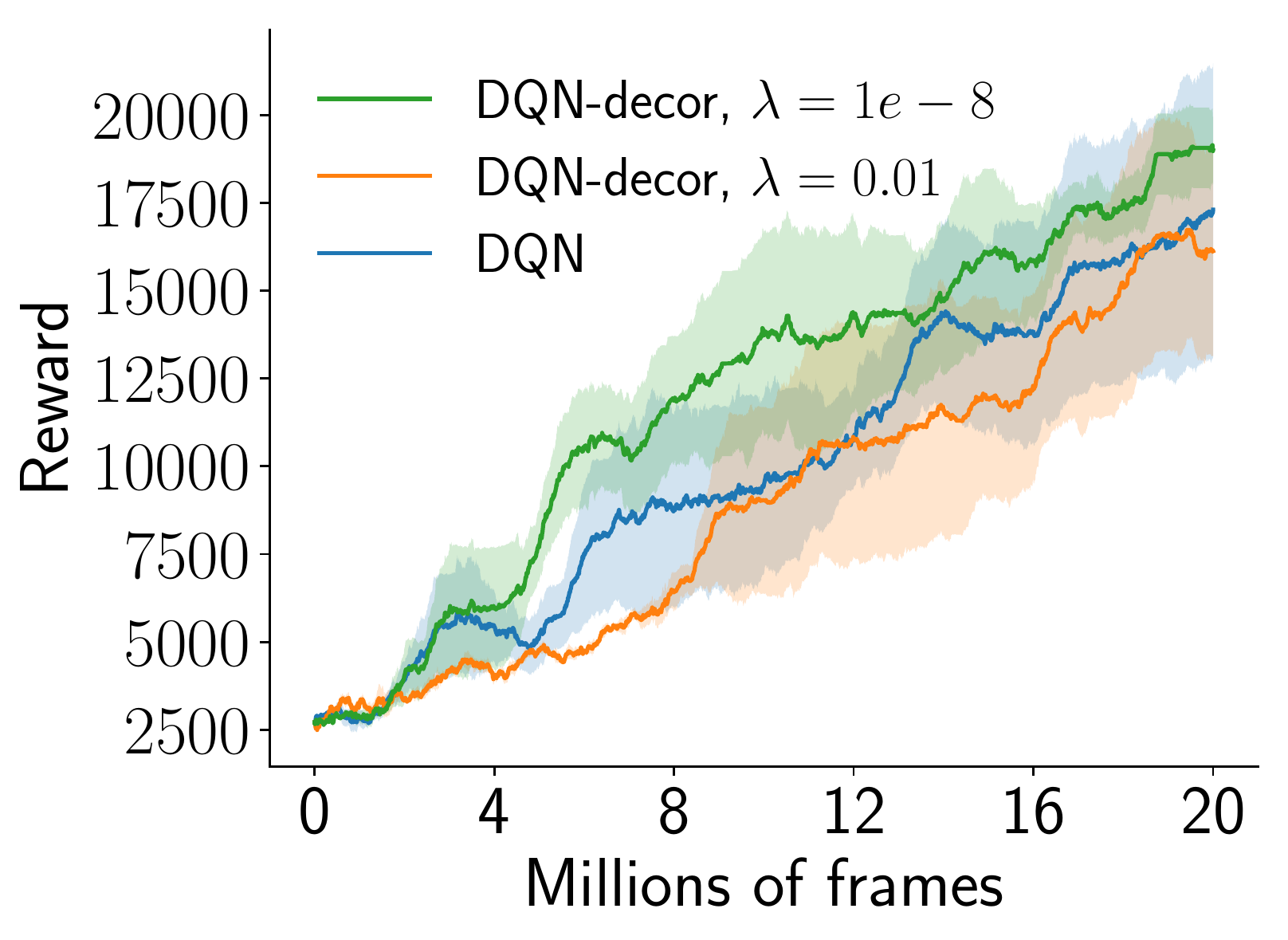}}
\caption{BattleZone: DQN-decor-$\lambda$-$10^{-8}$ performs significantly better than DQN. }
\label{fig:dqn-battlezone}
\end{subfigure}
\end{figure*}

The input of the neural networks is $(84,84,4)$ taking the most recent 4 frames. 
There are three convolution layers followed by a fully connected layer. 
The first layer convolves 32 filters of $8 \times 8$ with stride 4 with the input image and applies a rectifier non-linearity. 
The second layer convolves 64 filters of $4\times 4$ with strides 2, followed by a rectifier non-linearity. 
The third convolution layer convolves 64 filters of $3 \times 3$ with stride 1 followed by a rectifier.     
The final hidden layer has $512$ rectifier units. 
The outputs of these units given an image are just the features encoded in $\phi$. 
Thus the number of features, $d$, is $512$.  
This is followed by the output layer which is fully connected and linear, with a single output for each action (the number of actions in each game is different, ranging from 4 to 18). 

The $\kappa$ parameter for QR-DQN algorithms in the Huber loss function was set to $\kappa=1$ following \citet{dabney2017distributional}. 

The correlation loss in equation \ref{eqn:correlation-matrix} is computed on the minibatch sampled in experience replay (the empirical mean is also computed over the minibatch). 

\subsection{Decorrelating DQN}
\textbf{Performance}. First we compare DQN-decor with DQN in Figure \ref{fig:dqn-corr-perf}.  
The performance is measured by the median performance of DQN-decor and DQN across all games, where for each game the average score over runs is first taken. 
DQN-decor is shown to outperform DQN starting from about 4 million frames, with the winning edge increasing with time. 
By 20 million frames, DQN-decor achieves a human-normalized score of $70\%$ (while DQN is $50\%$), outperforming DQN by $40\%$.

While the median performance is widely used, it is only a performance summary across all games. 
To see how algorithms perform in each game,  we benchmark them for each game in Figure \ref{fig:dqn-auc}. 
DQN-decor won over DQN on $39$ games (each of which is more than $5\%$) with 4 close ties and lost slightly on $6$ games. 
In particular, only on Bowling and BattleZone, it loses to DQN by about $10\%$. 
For the other $4$ losing games, the loss was below $5\%$. 

We dig into the training on Bowling, and found that DQN fails to learn for this task. 
{The evaluation score for Bowling in the original DQN paper \cite{mnih2015human} is $42.4$ (see their Table 2) with a standard deviation of $88.0$ (trained over 50 million frames). 
In fact, for a fair comparison, one can check that the training performance for DQN at \url{https://google.github.io/dopamine/baselines/plots.html}, 
because scores reported in the original DQN paper is the testing performance without exploration. 
Their results show that for Bowling, the scores of DQN in the training phases fluctuate around $30.0$, which is close to our implemented DQN here. 
}
Figure \ref{fig:dqn-bowling-3runs} shows that two runs (out of three) of DQN is even worse than the random agent. 
Because the exploration factor is close to 1.0 in the beginning, the first data points of the curves correspond to the performance of the random agent. The representation learned by DQN turns out to be not useful for this task. 
In fact, the features learned by these two failure runs of DQN are nearly i.i.d, reflected in that the correlation loss is nearly zero as shown in Figure \ref{fig:dqn-bowling-3runs-cor}. 
So decorrelating the unsuccessful representation does not help in this case.

The loss of DQN-decor to DQN on BattleZone is due to the regularization factor, $\lambda$, which is uniformly fixed for all games. 
In producing these two figures, $\lambda$ was fixed to $0.01$ for all 49 games. 
In fact, we did not run parameter search to optimize $\lambda$ for all games. 
The value $\lambda=0.01$ was picked at our best guess. 
We did notice that DQN-decor can achieve better performance than DQN for BattleZone with a different regularization factor as shown in Figure \ref{fig:dqn-battlezone}. 
It appears that the features learned by DQN are already much decorrelated, and thus a small regularization factor (in this case, $10^{-8}$) gives a better performance.  

Excluding the unsuccessful representation learned by DQN for the case of Bowling and better parameter choice for regularization factor, 
we found that decorrelation is a uniformly effective technique for training DQN.

\textbf{Effect of Feature Correlation}. 
To study the correlation between performance and correlation loss, we plot the correlation loss over training time (also using the median measure) across 49 games in Figure \ref{fig:dqn-corr-loss}. The correlation loss is computed by averaging over the minibatch samples during experience replay at each time step. 

As shown by the figure, DQN-decor effectively reduces the decorrelation in features. We can see that the feature correlation loss for DQN-decor is almost flat from no more than 2 million frames, indicating that decorrelating features can be achieved faster than learning (a good news for representation learning). While for DQN, the learned features keep correlating with each other more and more in the first 6 million frames. After that, however, it appears that DQN does achieve a flat correlation loss, although much bigger than DQN-decor.


\begin{figure}[ht]
\vskip 0.2in
\begin{center}
\centerline{\includegraphics[width=\columnwidth]{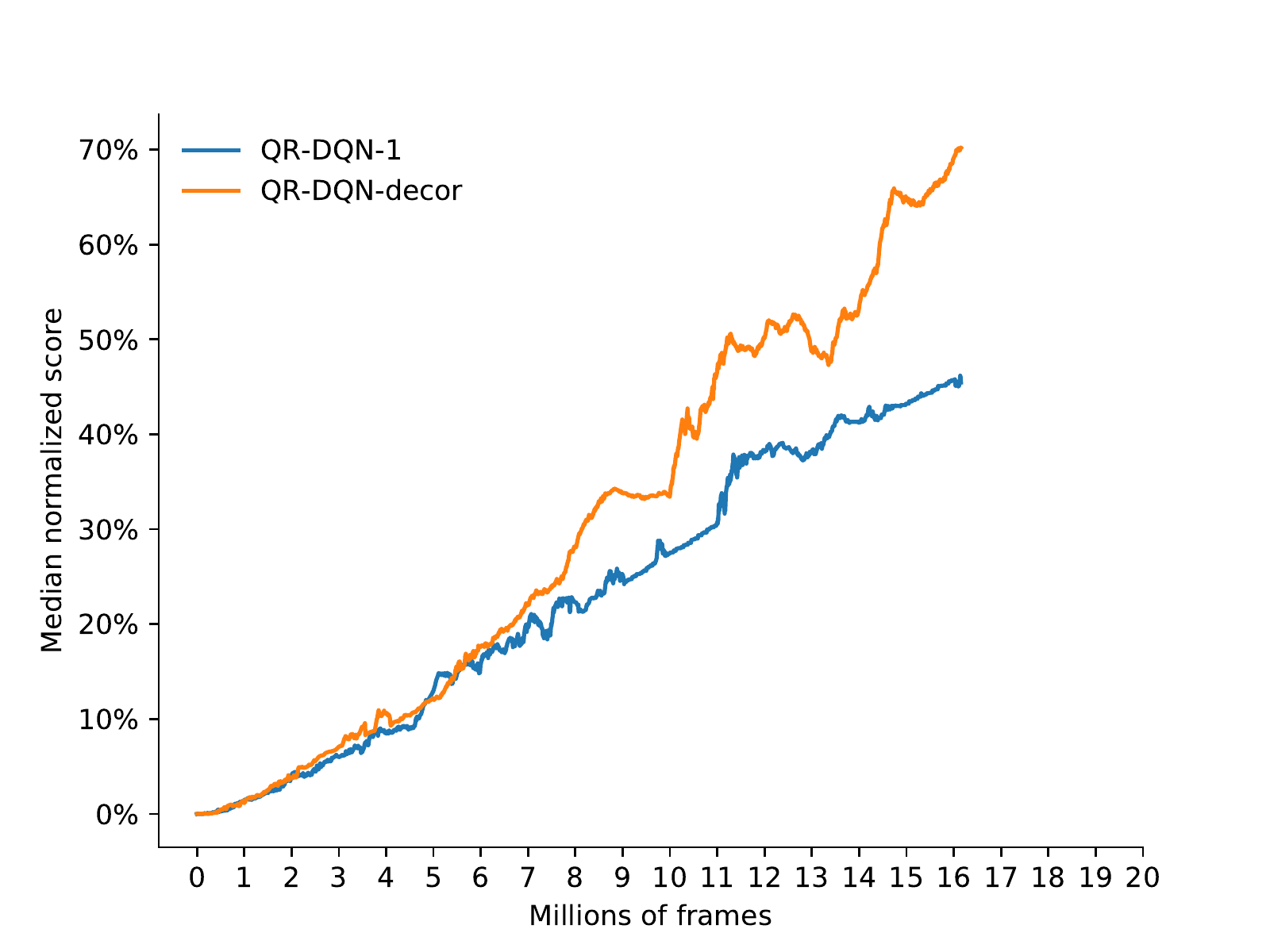}}
\caption{Human-normalized performance (using median across 49 Atari games): QR-DQN-decor vs. QR-DQN.}
\label{fig:qrdqn-corr-perf}
\end{center}
\vskip -0.2in
\end{figure}

\begin{figure}[ht]
\vskip 0.2in
\begin{center}
\centerline{\includegraphics[width=\columnwidth]{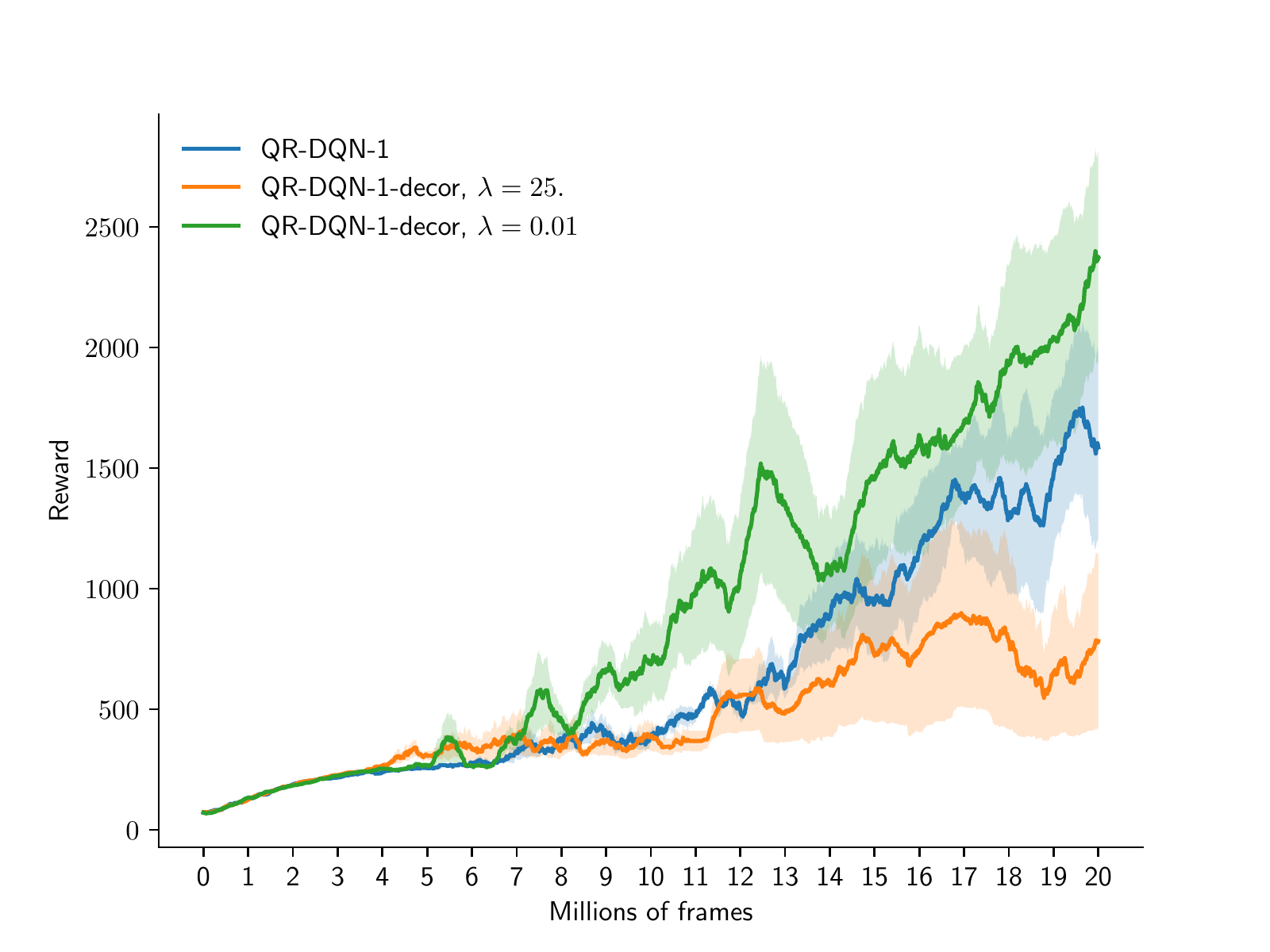}}
\caption{Frostbite: QR-DQN-decor-$\lambda$-0.01 won by $42.2\%$ in normalized AUC (area under curve). }
\label{fig:qrdqn-corr-perf-frostbite}
\end{center}
\vskip -0.2in
\end{figure}

\begin{figure*}[ht]
\vskip 0.2in
\begin{center}
\centerline{\includegraphics[width=1.5\columnwidth]{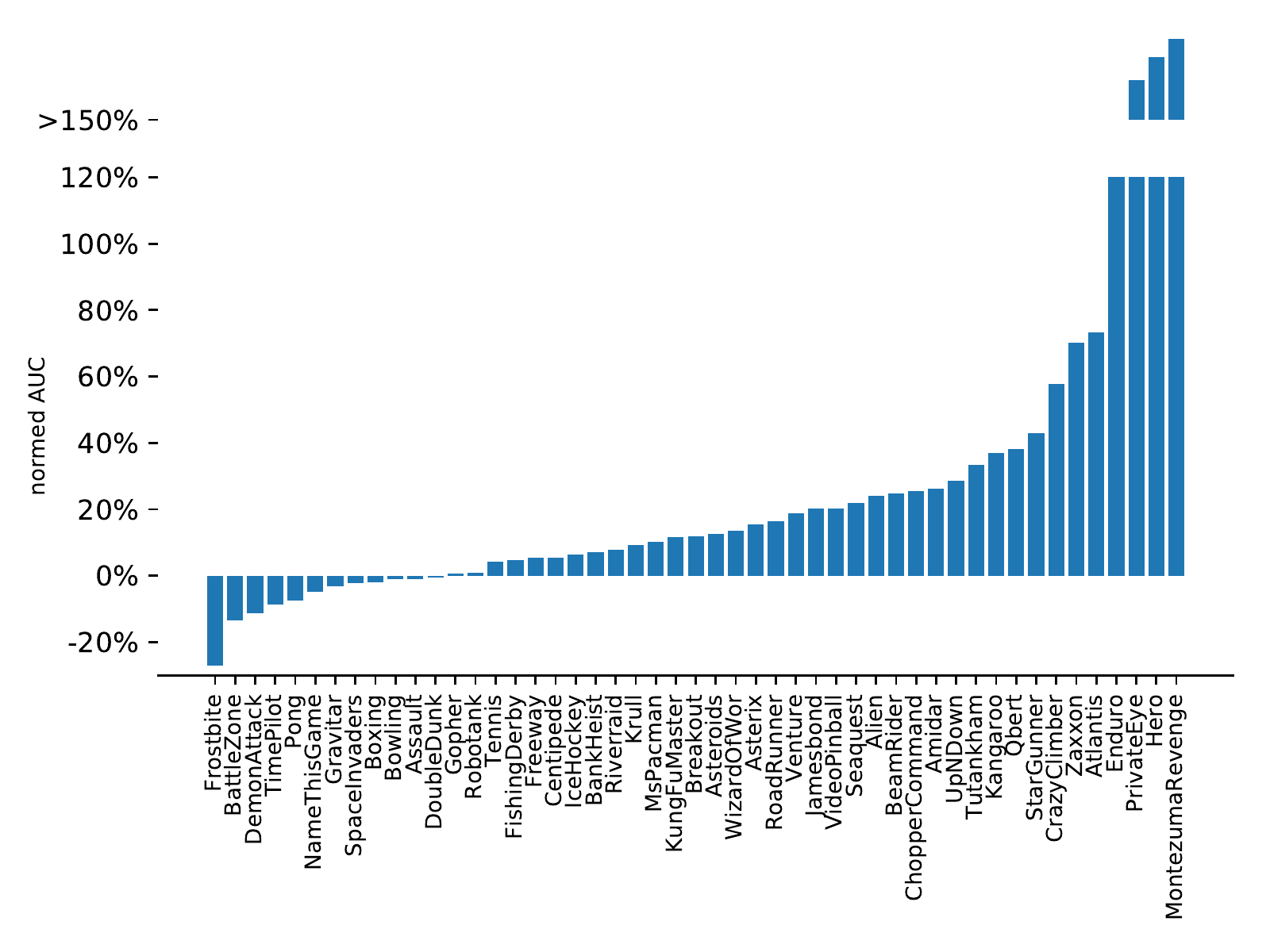}}
\caption{Normalized AUC: Cumulative reward improvement of QR-DQN-decor over QR-DQN.
For MonteZumaRevenge, the improvement is $419.31\%$; for Hero, the improvement is $414.22\%$; and for PrivateEye, the improvement is $318.18\%$.
}
\label{fig:qrdqn-auc}
\end{center}
\vskip -0.2in
\end{figure*}


\begin{figure}[t]
    \begin{subfigure}[t]{0.47\linewidth}        
        \centering
        \includegraphics[width=\linewidth]{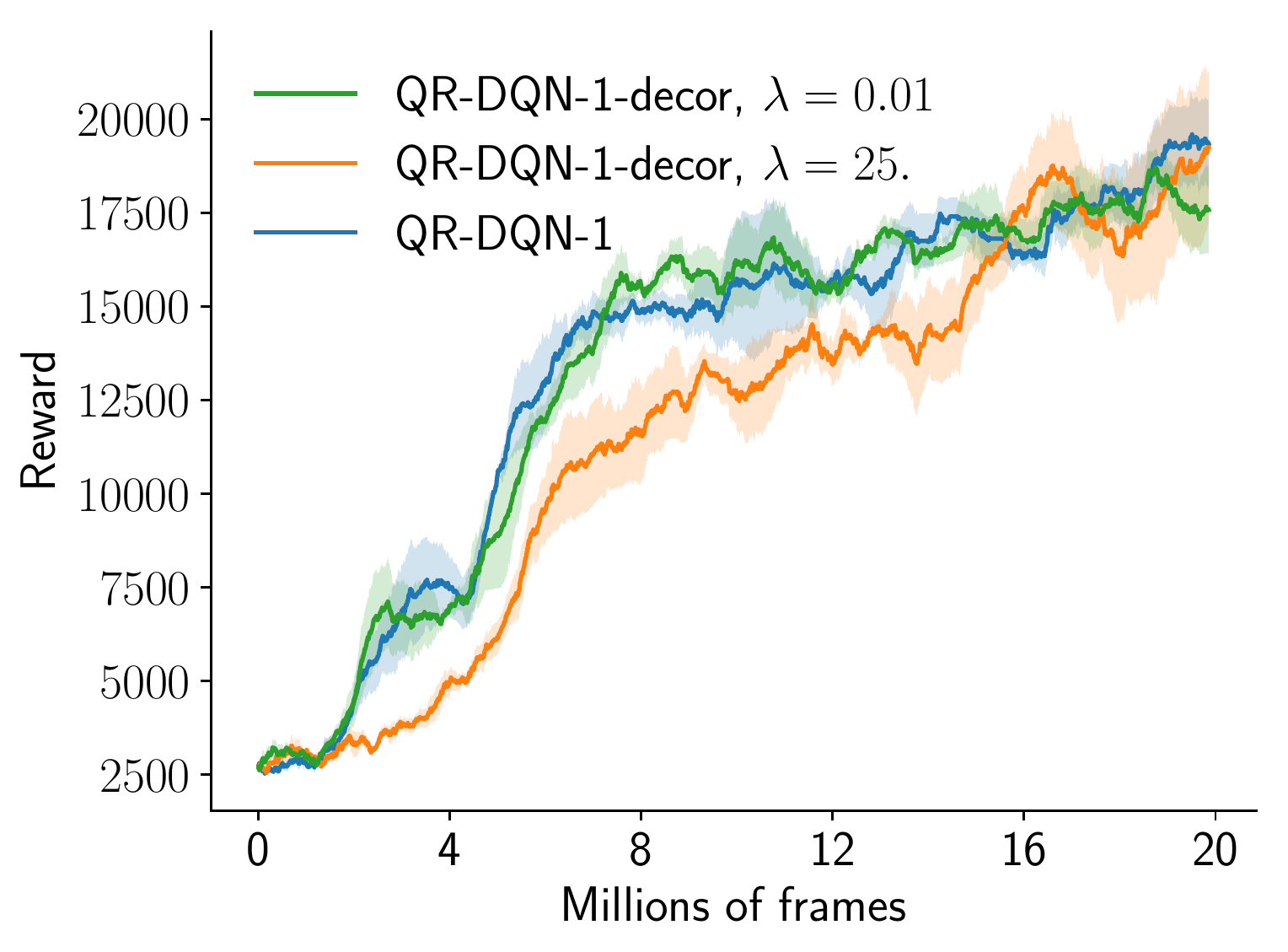}
        \caption{BattleZone.}
       \label{fig:qrdqn-corr-perf-battlezone}
    \end{subfigure}
	\begin{subfigure}[t]{0.47\linewidth}
	\centering
	\centerline{\includegraphics[width=\linewidth]{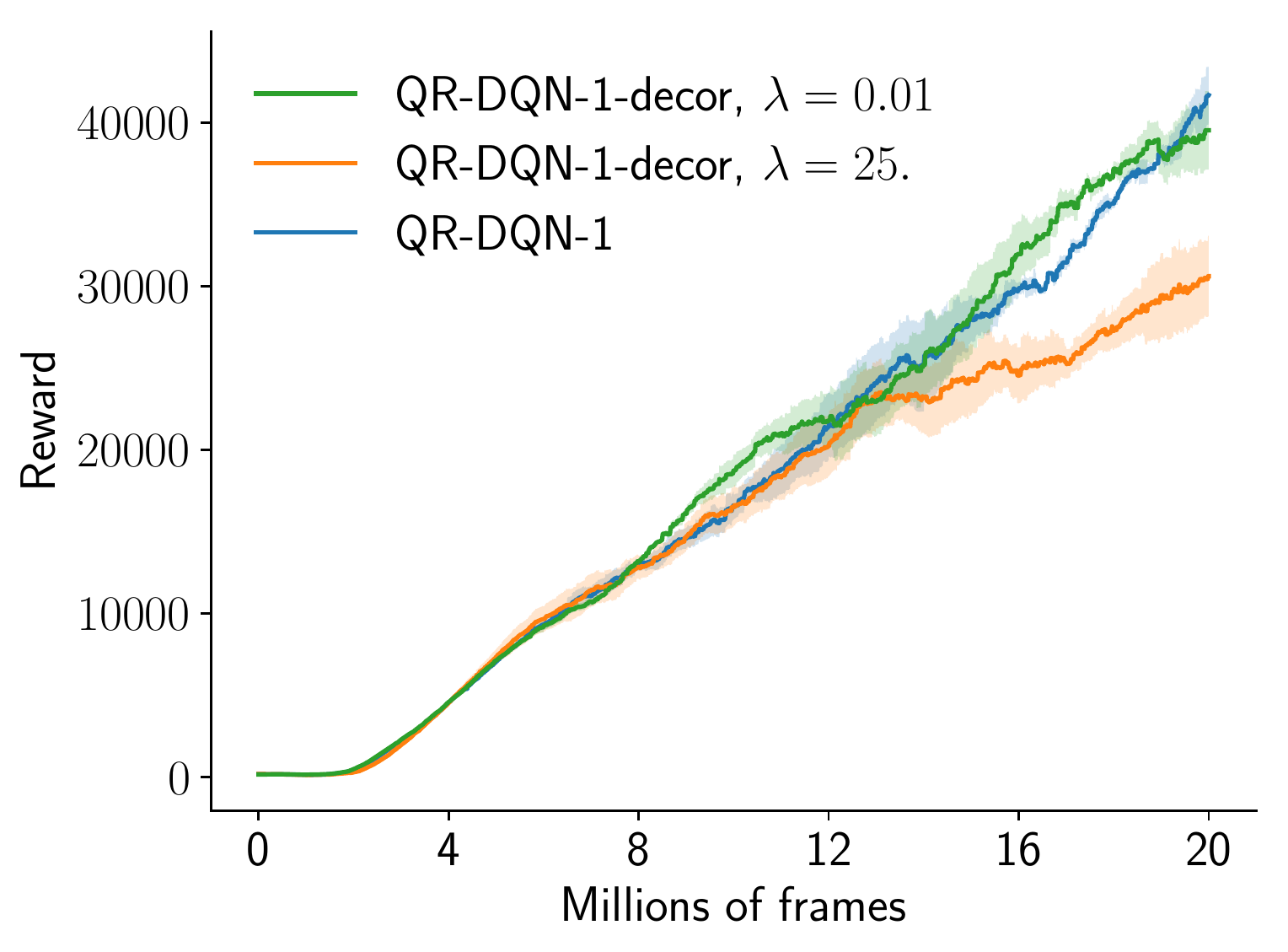}}
	\caption{DemonAttack.}
        \label{fig:qrdqn-corr-perf-DemonAttack}
	\end{subfigure}
        \caption{Re-comparing QR-DQN-decor (with $\lambda=0.01$) and QR-DQN in normalized AUC. Left (BattleZone): losing to QR-DQN by $-0.18\%$.
Right (DemonAttack): winning QR-DQN by $3.5\%$. 
}
\end{figure}

This phenomenon of DQN learning is interesting because it shows that although correlation is not considered in its loss function, DQN does have the ability of achieving feature decorrelation (to some extent) over time. 
Note that DQN's performance keeps improving after 6 million frames while at the same time its correlation loss measure is almost flat. 
This may indicate that it is natural to understand learning in two phases: decorrelating features (with some loss) and after that learning and performance continues to improve with decorrelated features.

Interestingly, faster learning of DQN-decor follows after the features are decorrelated. 
To see this, DQN-decor's advantage over DQN becomes significant after about 6 million frames (Figure \ref{fig:dqn-corr-perf}), 
while DQN-decor's correlation loss has been flat since less than 2 million frames (Figure \ref{fig:dqn-corr-loss}).

\begin{table*}[h]
\centering
\begin{tabular}{|l|r|r|r|r|}
\hline
\textbf{Game} & \textbf{DQN} & \textbf{DQN-decor, $\lambda=0.01$} & \textbf{QR-DQN-1} & \textbf{QR-DQN-1-decor, $\lambda=25$} \\
\hline
Alien & 1445.8 & 1348.5 & 1198.0 & \textbf{1496.0} \\
Amidar & 234.9 & 263.8 & 236.5 & \textbf{339.0} \\
Assault & 2052.6 & 1950.8 & \textbf{6118.2} & 5341.3 \\
Asterix & 3880.2 & 5541.7 & 6978.4 & \textbf{8418.6} \\
Asteroids & 733.2 & 1292.0 & 1442.4 & \textbf{1731.4} \\
Atlantis & 189008.6 & \textbf{307251.6} & 65875.3 & 148162.4 \\
BankHeist & 568.4 & 648.1 & 730.3 & \textbf{740.8} \\
BattleZone & 15732.2 & 14945.3 & \textbf{18001.3} & 17852.6 \\
BeamRider & 5193.1 & 5394.4 & 5723.6 & \textbf{6483.1} \\
Bowling & \textbf{27.3} & 21.9 & 22.5 & 22.2 \\
Boxing & 85.6 & 86.0 & \textbf{87.1} & 83.9 \\
Breakout & 311.3 & 337.7 & 372.8 & \textbf{393.3} \\
Centipede & 2161.2 & 2360.5 & 6003.0 & \textbf{6092.9} \\
ChopperCommand & 1362.4 & 1735.2 & 2266.9 & \textbf{2777.4} \\
CrazyClimber & 69023.8 & \textbf{100318.4} & 74110.1 & 100278.4 \\
DemonAttack & 7679.6 & 7471.8 & \textbf{34845.7} & 27393.6 \\
DoubleDunk & \textbf{-15.5} & -16.8 & -18.5 & -19.1 \\
Enduro & 808.3 & \textbf{891.7} & 409.4 & 884.5 \\
FishingDerby & 0.7 & \textbf{11.7} & 9.0 & 10.8 \\
Freeway & 23.0 & \textbf{32.4} & 25.6 & 24.9 \\
Frostbite & 293.8 & 376.6 & \textbf{1414.2} & 755.7 \\
Gopher & 2064.5 & 3067.6 & 2816.5 & \textbf{3451.8} \\
Gravitar & 271.2 & \textbf{382.3} & 305.9 & 314.9 \\
Hero & 3025.4 & 6197.1 & 1948.4 & \textbf{9352.2} \\
IceHockey & -10.0 & \textbf{-8.6} & -10.3 & -9.6 \\
Jamesbond & 387.5 & 471.0 & 391.4 & \textbf{515.0} \\
Kangaroo & 3933.3 & \textbf{3955.5} & 1987.8 & 2504.4 \\
Krull & 5709.9 & 6286.4 & 6547.7 & \textbf{6567.9} \\
KungFuMaster & 16999.0 & 20482.9 & 22131.3 & \textbf{23531.4} \\
MontezumaRevenge & 0.0 & 0.0 & 0.0 & \textbf{2.4} \\
MsPacman & 2019.0 & 2166.0 & 2221.4 & \textbf{2407.9} \\
NameThisGame & 7699.0 & 7578.2 & \textbf{8407.0} & 8341.0 \\
Pong & 19.9 & \textbf{20.0} & 19.9 & 20.0 \\
PrivateEye & 345.6 & \textbf{610.8} & 41.7 & 251.0 \\
Qbert & 2823.5 & 4432.4 & 4041.0 & \textbf{5148.1} \\
Riverraid & 6431.3 & 7613.8 & 7134.8 & \textbf{7700.5} \\
RoadRunner & 35898.6 & \textbf{39327.0} & 36800.0 & 37917.6 \\
Robotank & 24.8 & 24.5 & \textbf{31.3} & 30.6 \\
Seaquest & 4216.6 & \textbf{6635.7} & 4856.8 & 5224.6 \\
SpaceInvaders & \textbf{1015.8} & 913.0 & 946.7 & 825.1 \\
StarGunner & 15586.6 & 21825.0 & 25530.8 & \textbf{37461.0} \\
Tennis & -22.3 & -21.2 & -17.3 & \textbf{-16.2} \\
TimePilot & 2802.8 & \textbf{3852.1} & 3655.8 & 3651.6 \\
Tutankham & 103.4 & 116.2 & 148.3 & \textbf{190.2} \\
UpNDown & 8234.5 & 9105.8 & 8647.4 & \textbf{11342.4} \\
Venture & 8.4 & \textbf{15.3} & 0.7 & 1.1 \\
VideoPinball & 11564.1 & 15759.3 & 53207.5 & \textbf{66439.4} \\
WizardOfWor & 1804.3 & 2030.3 & 2109.8 & \textbf{2530.6} \\
Zaxxon & 3105.2 & \textbf{7049.4} & 4179.8 & 6816.9 \\
\hline
\end{tabular}
\caption{Comparison of gaming scores obtained by our decorrelation algorithms with DQN and QR-DQN, averaged over the last one million training frames (the total number training frames is 20 million). 
Out of 49 games, decorrelation algorithms perform the best in 39 games: DQN-decor is best in 15 games and QR-DQN-decor is best in 24 games. 
}
\label{table:atari}
\end{table*}

\subsection{Decorrelating QR-DQN}
We further conduct experiments to study whether our decorrelation method applies to QR-DQN. 
The value of $\lambda$ was set to $25.0$, picked simply from a parameter study on the game of Sequest. 
Figure \ref{fig:qrdqn-corr-perf} shows the median human-normalized performance scores across 49 games.
Similar to the case of DQN, our method achieves much better performance by decorrelating QR-DQN, especially after about 7 million frames. 
Again, we see the performance edge of QR-DQN-decor over QR-DQN has a trend of increasing with time. 
We observed a similar correlation between performance and correlation loss, which supports that decorrelation is the factor that improves the performance. 

We also profile the performance of algorithms for each game in the normalized AUC, which is shown in Figure \ref{fig:qrdqn-auc}. 
In this case, the decorrelation algorithm lost 5 games, with 10 games being close (less than $5\%$), and 34 winning games (bigger than $5\%$). 
The algorithm lost most to QR-DQN on the game of Frostbite, BattleZone and DemonAttack. 
This is due to parameter choice instead of an algorithmic issue. 
For all the three losing games, we performed additional experiments with $\lambda = 0.01$ (the same value as for the DQN experiments). 
The learning curves are shown in Figure \ref{fig:qrdqn-corr-perf-frostbite} (Frostbite), Figure \ref{fig:qrdqn-corr-perf-battlezone} (BattleZone), and Figure \ref{fig:qrdqn-corr-perf-DemonAttack} (DemonAttack). 
Recall that QR-DQN-decor with $\lambda = 25.0$ lost to QR-DQN by about $25\%$. 
Figure \ref{fig:qrdqn-corr-perf-frostbite} shows that with $\lambda=0.01$, the performance of QR-DQN-decor is significantly improved, 
with an improvement over QR-DQN by $42.2\%$ (in the normalized AUC measure).  
For the other two games, QR-DQN-decor performs at least no worse than QR-DQN ($3.5\%$ better on DemonAttack, $-0.18\%$ on BattleZone). 

Note that the median human-normalized scores across games and the normalized AUC measure for each game may not give a full picture of algorithm performances. 
Algorithms perform well in these two measures can exhibit plummeting behaviour. 
Plummeting is characterized by abrupt degradation in performance. 
In this case the learning curve can drop to low score and stay there indefinitely. A more detailed discussion of this point is in \cite{machado2017revisiting}.
To study if our decorrelation algorithms have plummeting behaviour, we benchmark all algorithms by averaging the rewards in the last one million training frames (which is essentially near-tail performance in training). 
In this measure, the results are summarized in Table \ref{table:atari}. 
Out of 49 games, decorrelation algorithms perform the best in 39 games: DQN-decor is best in 15 games and QR-DQN-decor is best in 24 games. 
Thus our decorrelation algorithms do not have plummeting behaviour. 

In a summary, empirical results in this section show that decorrelation is effective for improving the performance of both QR-DQN and QR-DQN. 
Even on the most losing games (incurred using a fixed, same regularization factor for all games), both DQN and QR-DQN can still be significantly improved by our decorrelation method with a tuned regularization factor. 

\subsection{Analysis on the Learned Representation}
To study the correlation between the learned features after training finished, 
we first compute the feature covariance matrix (see equation \ref{eqn:correlation-matrix}) using the (one million) samples in the experience replay buffer at the end of training, for both DQN and DQN-decor. 
Then we sort the (512) features according to their variances, which are just the diagonal parts of the feature covariance matrix.

\begin{figure}[t]
    \begin{subfigure}[b]{0.47\linewidth}        
        \centering
        \includegraphics[width=\linewidth]{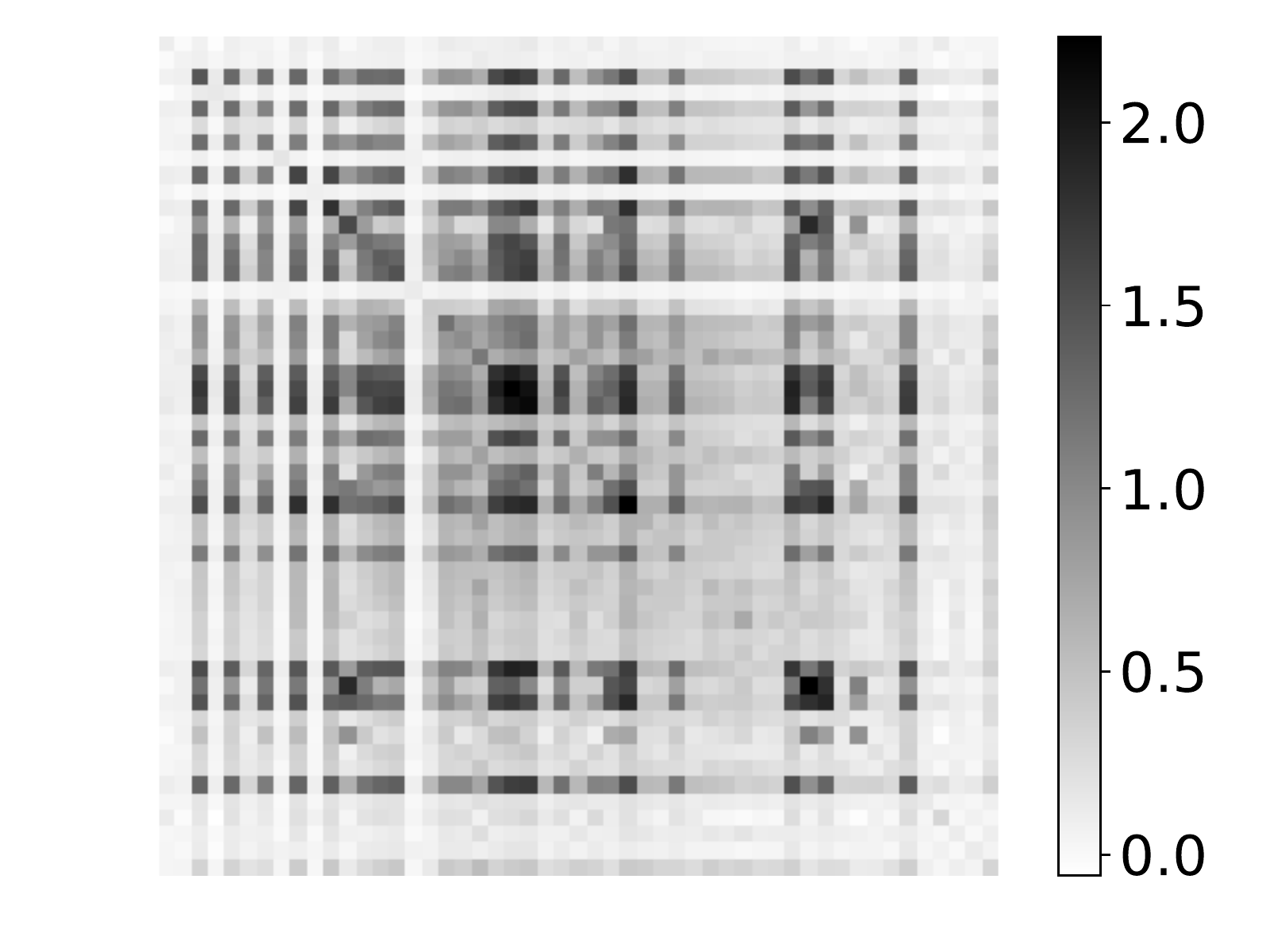}
        \caption{DQN.}
        \label{fig:dqn-features-sequest}
    \end{subfigure}
	\begin{subfigure}[b]{0.47\linewidth}
	\centering
	\centerline{\includegraphics[width=\columnwidth]{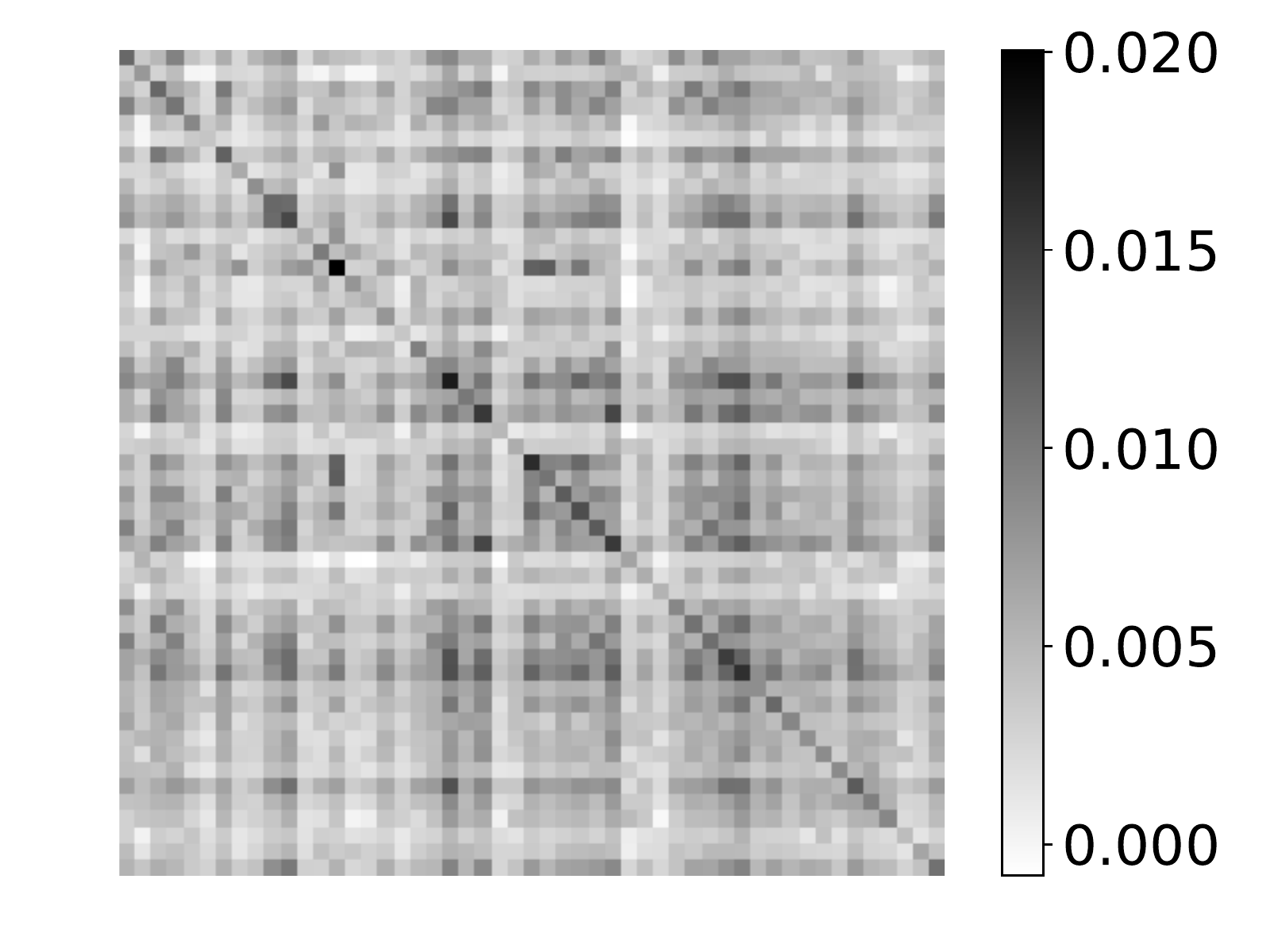}}
	        \caption{DQN-decor. }
	\label{fig:dqndec-features-sequest}
	\end{subfigure}
    \caption{Features correlation heat map on Sequest. First, the diagonalization pattern is more obvious in the feature covariance matrix of DQN-decor. 
Second, the magnititude of feature correlation in DQN is much larger than DQN-decor. Third, there are more highly activated features in DQN and they correlate with each other, suggesting that many of them may be redundent. }
\end{figure}

\begin{figure*}
\centering
  \includegraphics[width=\linewidth]{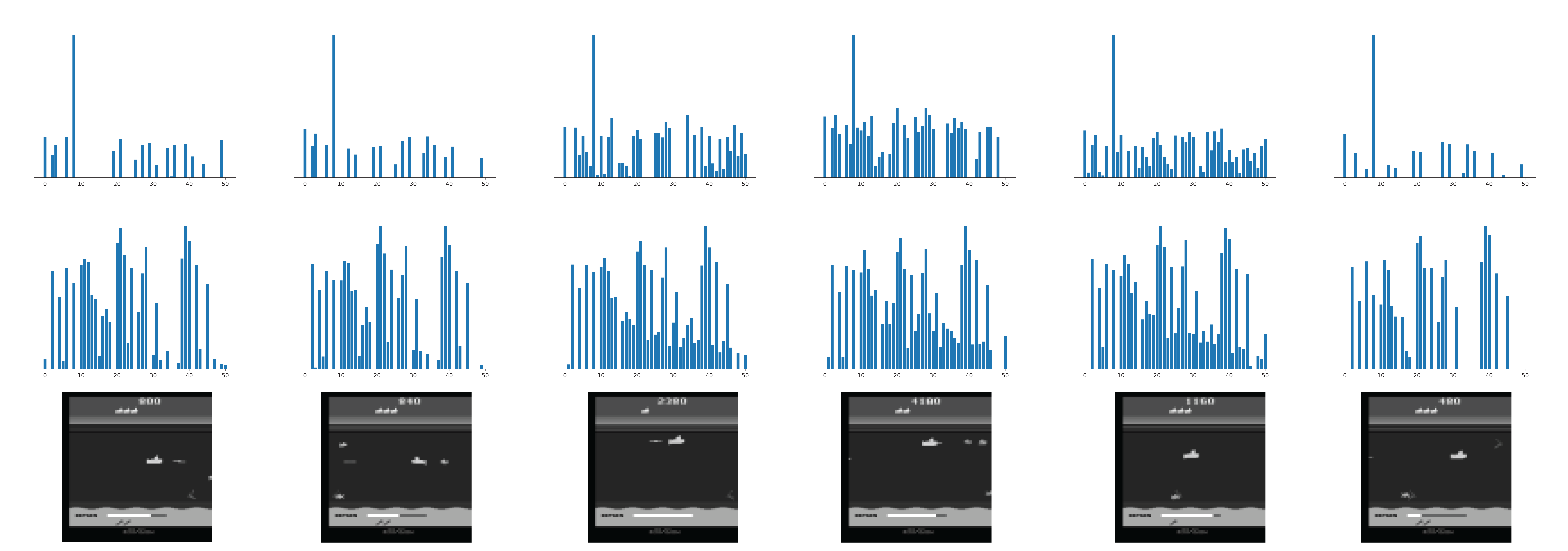}
 \caption{Activation of the most important features (according to their variances) on Sequest (6 random samples). Top row: features of DQN-dec. Middle row: features of DQN. Bottom row: current image frame. }
  \label{fig:features-sequest}
\end{figure*}

A visualization of the top 50 features on the game of Sequest is shown in Figure \ref{fig:dqn-features-sequest} (DQN), and Figure \ref{fig:dqndec-features-sequest} (DQN-decor). Three observations can be made. First, the diagonalization pattern is more obvious in the feature covariance matrix of DQN-decor. Second, the magnititude of feature correlation in DQN is much larger than DQN-decor (note the heat intensity bar on the right for the two plots is different). Third, there are more highly activated features in DQN, reflected in that the diagonal parts of DQN's matrix have more intense values (black color). Interestingly, there are also almost same number of equally sized squares in the off-diagonal region of DQN's matrix, indicating that these highly activated features correlate with each other and many of them may be redundent. In contrast, the off-diagonal region of DQN-decor's matrix has very few intense values, suggesting that these important features are successfully decorrelated.    

Figure \ref{fig:features-sequest} shows randomly sampled frames and their feature activation values (same set of 50 features as in Figure \ref{fig:dqn-features-sequest} and Figure \ref{fig:dqndec-features-sequest}) for both algorithms. 
Interestingly, the features of DQN-decor (top row) appear to have much fewer values that are activated at the same time given an input image. 
In addition, DQN also has many highly activated features at the same time. In contrast, for DQN-decor, there are just a few (in these samples, there is only one) highly active features. 
Interpretting features for DRL algorithms may be made easier thanks to decorrelation. 
Although we have not any conclusion on this yet, it is an interesting future research direction. 

\section{Conclusion}
In this paper, we found that feature correlation is a key factor in the performance of DQN algorithms. 
We have proposed a method for obtaining decorrelated features for DQN.
The key idea is to regularize the loss function of DQN by a correlation loss computed from the mean squared correlation between the  features. 
Our decorrelation method turns out to be very effective for training DQN algorithms. 
We show that it also applies to QR-DQN, improving QR-DQN significantly on most games, and losing only on a few games. 
Experiment study on the losing games shows that it is due to the parameter choice for the regularization factor. 
Our decorrelation method effectively improves DQN and QR-DQN with a better or no worse performance across most games, 
which makes it hopeful for improving representation with the other DRL algorithms. 


\appendix

\section{All Learning Curves}
\begin{figure*}[t]
\vskip 0.2in
\begin{center}
\centerline{\includegraphics[width=2\columnwidth]{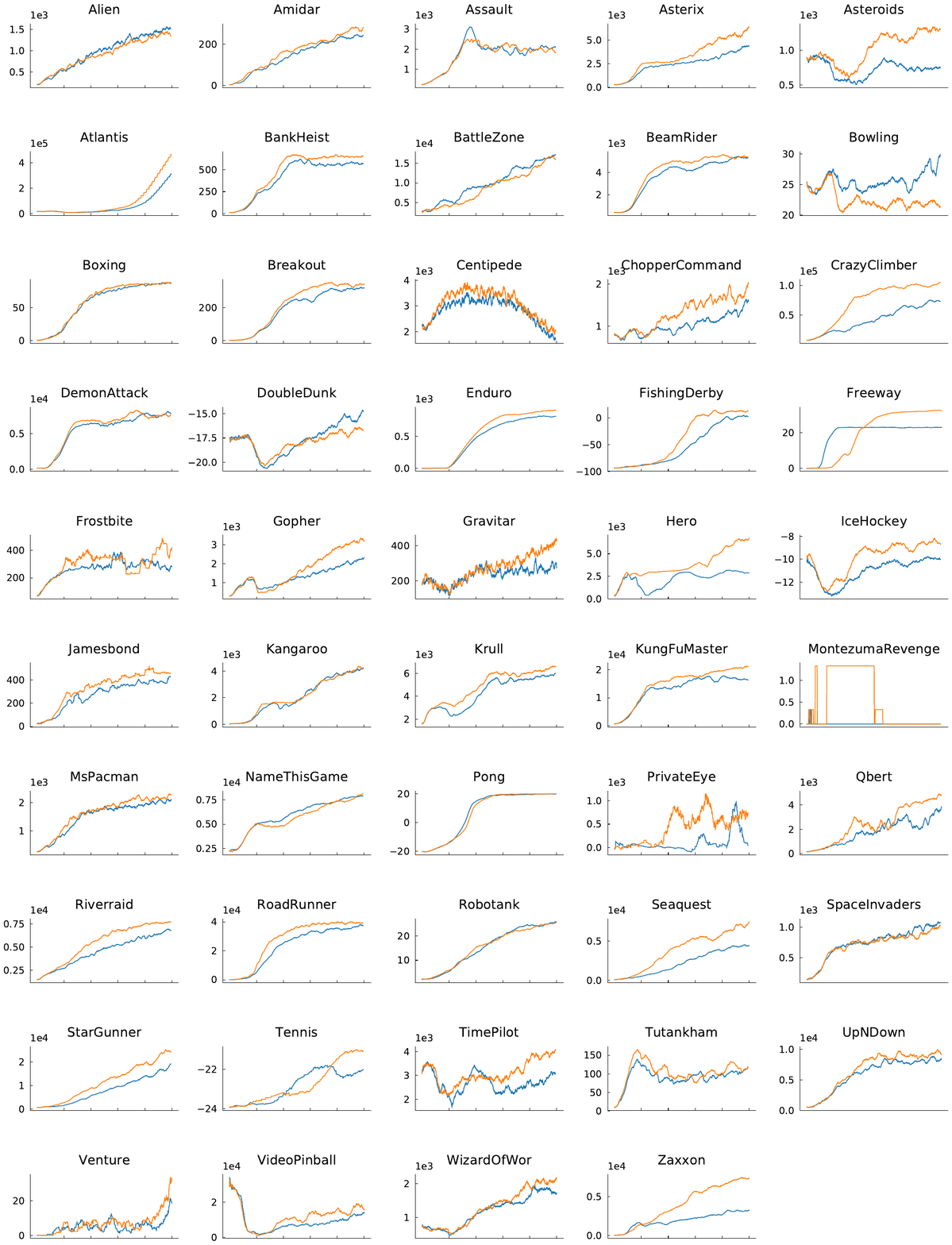}}
\caption{DQN-decor (\textcolor{orange}{orange}) vs.  DQN (\textcolor{blue}{blue}): all learning curves for 49 Atari games. }
\label{fig:dqn-all}
\end{center}
\vskip -0.2in
\end{figure*}

The learning curves of DQN-decor vs. DQN on all games are shown in Figure \ref{fig:dqn-all}. ($\lambda=0.01$). 

\begin{figure*}[t]
\vskip 0.2in
\begin{center}
\centerline{\includegraphics[width=2\columnwidth]{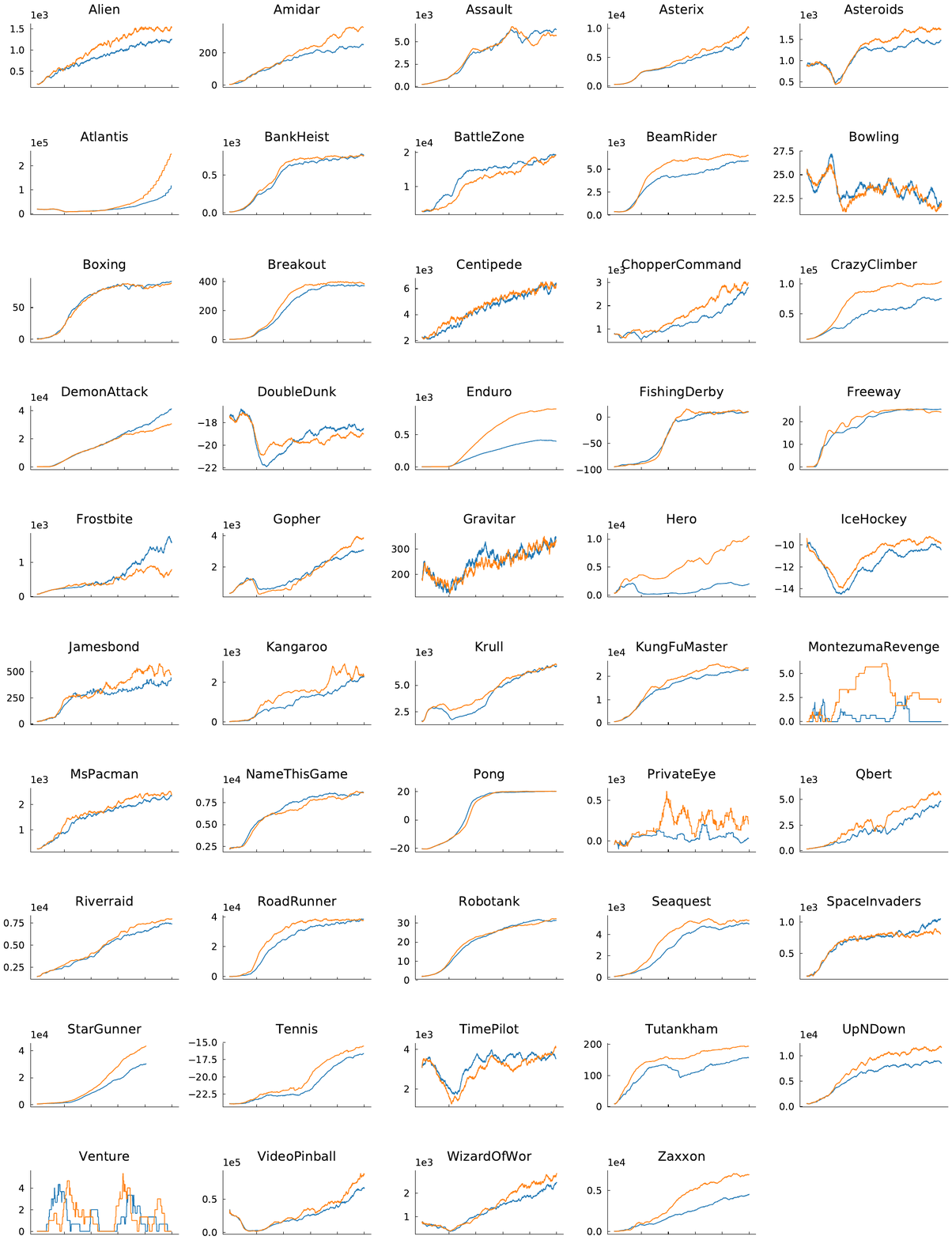}}
\caption{QR-DQN-decor (\textcolor{orange}{orange}) vs. QR-DQN (\textcolor{blue}{blue}): all learning curves for 49 Atari games. }
\label{fig:qrdqn-all}
\end{center}
\vskip -0.2in
\end{figure*}

The learning curves of QR-DQN-decor vs. QR-DQN are shown in Figure \ref{fig:qrdqn-all}. ($\lambda=25.0$). 

\bibliography{/home/hengshuai/reference}

\bibliographystyle{icml2019}

\end{document}